\documentclass[conference]{IEEEtran}
\IEEEoverridecommandlockouts
\usepackage{cite}
\usepackage{amsmath,amssymb,amsfonts}
\usepackage{algorithmic}
\usepackage{graphicx}
\usepackage{textcomp}
\usepackage{xcolor}
\def\BibTeX{{\rm B\kern-.05em{\sc i\kern-.025em b}\kern-.08em
    T\kern-.1667em\lower.7ex\hbox{E}\kern-.125emX}}
\usepackage[colorlinks=false]{hyperref}
\begin{document}

\title{Analysis Towards Classification of Infection and Ischaemia of Diabetic Foot Ulcers}

\author{\IEEEauthorblockN{Moi Hoon Yap\IEEEauthorrefmark{1}, 
Bill Cassidy\IEEEauthorrefmark{1}, 
Joseph M. Pappachan\IEEEauthorrefmark{2}\IEEEauthorrefmark{1},
Claire O'Shea\IEEEauthorrefmark{3},
David Gillespie\IEEEauthorrefmark{1} and 
Neil D. Reeves\IEEEauthorrefmark{1}}
\IEEEauthorblockA{\IEEEauthorrefmark{1}Faculty of Science and Engineering, Manchester Metropolitan University, Manchester, UK}
\IEEEauthorblockA{\IEEEauthorrefmark{2}Lanchashire Teaching Hospitals, UK}
\IEEEauthorblockA{\IEEEauthorrefmark{3}Waikato District Health Board, Hamilton, New Zealand } }

\maketitle

\begin{abstract}
This paper introduces the Diabetic Foot Ulcers dataset (DFUC2021) for analysis of pathology, focusing on infection and ischaemia. We describe the data preparation of DFUC2021 for ground truth annotation, data curation and data analysis. The final release of DFUC2021 consists of 15,683 DFU patches, with 5,955 training, 5,734 for testing and 3,994 unlabeled DFU patches. The ground truth labels are four classes, i.e. control, infection, ischaemia and both conditions. We curate the dataset using image hashing techniques and analyse the separability using UMAP projection. We benchmark the performance of five key backbones of deep learning, i.e. VGG16, ResNet101, InceptionV3, DenseNet121 and EfficientNet on DFUC2021. We report the optimised results of these key backbones with different strategies. Based on our observations, we conclude that EfficientNetB0 with data augmentation and transfer learning provided the best results for multi-class (4-class) classification with macro-average Precision, Recall and F1-Score of 0.57, 0.62 and 0.55, respectively. In ischaemia and infection recognition, when trained on one-versus-all, EfficientNetB0 achieved comparable results with the state of the art. Finally, we interpret the results with statistical analysis and Grad-CAM visualisation.
\end{abstract}


\section{Introduction}
Diabetes is a global epidemic affecting ~425 million people and expected to rise to 629 million people by 2045 \cite{cho2018idf}. Diabetic Foot Ulcers (DFU) are a serious complication of diabetes affecting one in three people with diabetes \cite{armstrong2017diabetic}. The rapid rise in the prevalence of DFUs over the last few decades is a major challenge for healthcare systems around the world. DFU with infection and ischaemia can significantly prolong treatment, and often result in limb amputation, with more serious cases leading even to death. In an effort to improve patient care and reduce the strain on healthcare systems, early detection of DFU and regular monitoring by patients themselves (or a carer/partner) are important. Recent research has focused on the creation of detection algorithms that could be used as part of a mobile app that empower patients (or a carer/partner) in this regard \cite{yap2018new, goyal2018robust}.

The collaborative work between Manchester Metropolitan University, Lancashire Teaching Hospitals has created a repository of DFU images with clinically annotated infection and ischaemia cases for the purpose of supporting research toward more advanced methods of DFU pathology recognition \cite{goyal2018dfunet,goyal2020recognition}. Following the success of DFUC2020 \cite{yap2020deep}, lead scientists from the UK, US, India and New Zealand launched a joint effort to conduct Diabetic Foot Ulcers Grand Challenge (DFUC) 2021 to continue to solicit the original works in DFU, promote interactions between researchers and interdisciplinary collaborations. This paper describes the creation of the DFUC2021 dataset (henceforth DFUC2021), analyse the distribution of the dataset according to its pathology classes, benchmark the performance of deep learning classification frameworks on multi-class (4 classes) classification, interpret the classification results with Unified Uniform Manifold Approximation  and  Projection(UMAP) \cite{mcinnes2018umap}, statistics metrics and Grad-CAM visualisation \cite{selvaraju2017grad}.

\section{Methodology}
This section introduces DFUC2021 dataset, its ground truth labelling, pre-processing and preliminary analysis of DFUC2021, selection of deep learning backbones to benchmark the pathology classification and the performance metrics used to rank the results of the computer algorithms.

\subsection{Datasets and Ground Truth}
We have received approval from the UK National Health Service Research Ethics Committee (reference number is 15/NW/0539) to use diabetic foot ulcer (DFU) images for the purpose of this research. These images are photographs collected from the Lancashire Teaching Hospitals, where photographs were acquired from the patients during their clinical visits. Three cameras were used for capturing the foot images, Kodak DX4530, Nikon D3300 and Nikon COOLPIX P100. The images were acquired with close-ups of the full foot at a distance of around 30–40 cm with the parallel orientation to the plane of an ulcer. The use of flash as the primary light source was avoided, and instead, adequate room lights were used to get the consistent colours in image. A podiatrist and a consultant physician with specialisation in the diabetic foot (both with more than 5 years professional experience) helped to extract the images from the archive. 


\subsubsection{Ground Truth Annotation} The instruction for annotation is to identify the location of the ulcer with a bounding box and label each ulcer with ischaemia and/or infection, or none. Similar to DFUC2020 \cite{cassidy2020dfuc2020}, we use the software, LabelImg \cite{tzutalingit}, to label the images. There are multiple annotations from a podiatrist and a consultant physician with specialisation in the diabetic foot. We average the bounding boxes to form a final bounding box. For diabetic foot pathology classification, the pathology labels were validated with medical records.

\subsubsection{Data Curation}
\begin{figure}[!htb]
\centering
\includegraphics[width=.12\textwidth]{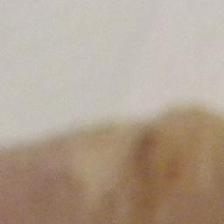}
\includegraphics[width=.12\textwidth]{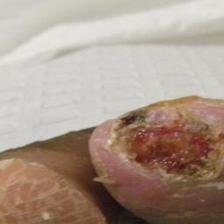}
\caption{An example of two different DFU images were grouped into the same cluster with average hashing. }
\label{figure:average}
\end{figure}

\begin{figure}[!htb]
\centering
\includegraphics[width=.12\textwidth]{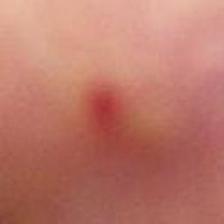}
\includegraphics[width=.12\textwidth]{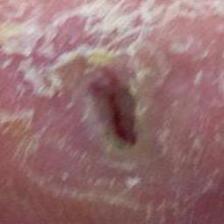}
\includegraphics[width=.12\textwidth]{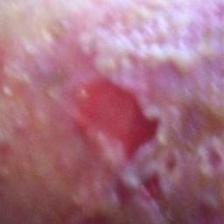}
\includegraphics[width=.12\textwidth]{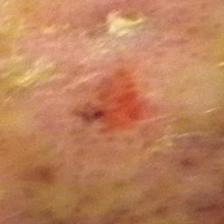}
\caption{An example of four different DFU images were grouped into the same cluster with wavelet hashing. }
\label{figure:wavelet}
\end{figure}

We cropped the DFU regions (from the final bounding box) and performed natural data augmentation (preserved case id and used it to split exclusive train and test set). We ran a sanity check to reduce the repetition of similar images. We tested image hashing techniques \cite{Buchner2020}, i.e. different hashing, average hashing, perceptual hashing and wavelets hashing to check the similarity of the DFU images. Different hashing finds the exact match of the images and perceptual hashing finds the images with different scales.  We have discarded images with exact matched and minor scaling changes (generated by natural data augmentation). Average hashing and wavelets hashing as shown in Figure \ref{figure:average} and Figure \ref{figure:wavelet} are not suitable for sanity check, as they clustered images with inter-similarities. These represent great challenges for the machine algorithms to automated the classification process.

For DFU classification, we followed a 50-50 split for training and testing set (case exclusion). From the training set, we further split 20\% from training set, which we used as a validation set. The total images for DFUC2021 is 15,683, with 5,955 images for training set (2,555 infection only, 227 ischaemia only, 621 both infection and ischaemia, and 2,552 without ischaemia and infection), 3,994 unlabeled images and 5,734 images for testing set. To obtain the dataset, visit: https://dfu-challenge.github.io

\begin{table*}
	\centering
	\renewcommand{\arraystretch}{1.0}
	\caption{A Comparison of the overall performance of the state-of-the-art methods with and without pretrained model, results reported on their best epoch.}
	\scalebox{1.0}{
	    \label{table:multi-class}
		\begin{tabular}{|l|cc|cccc|cc|cccc|}
			\hline
			Method  & \multicolumn{2}{l}{Settings}  &\multicolumn{10}{l|}{Metrics} \\
	               &&& \multicolumn{4}{l}{ Per class F1-Score} & \multicolumn{2}{l}{micro-average} & \multicolumn{4}{l|}{macro-average}\\
			
		 &Pretrained & Best epoch& Control & Infection & Ischaemia & Both & F1 & AUC & Precision & Recall & F1 & AUC\\
			\hline
			\hline
			VGG16 &$\times$&45&0.68& 0.51 & 0.44 & 0.43 &0.58& 0.86 & 0.50 & 0.57 & 0.51&0.83\\
			VGG16&\checkmark&19& 0.70 & 0.51 & 0.36 &0.43& 0.59 & 0.86& 0.49 & 0.54&0.50&0.83\\
			ResNet101&$\times$ &15&0.70 &0.45 & 0.42 & \textbf{0.52} & 0.59&0.86 & 0.54 & 0.57 & 0.52& 0.85\\
			ResNet101&\checkmark&24&0.71 &0.47 & 0.39 & 0.45 & 0.59&0.85 & 0.53 & 0.57 & 0.51& 0.84\\
			InceptionV3&$\times$ &13&0.73 &0.53 & 0.40 & 0.47 & 0.62&\textbf{0.87} & 0.53 & 0.58 & 0.53& 0.84\\
			InceptionV3&\checkmark&21&0.72 &0.54 & 0.42 & 0.45 & 0.62&0.86 & 0.53 & 0.57 & 0.53& 0.84\\
			DenseNet121 &$\times$ &17&0.69& 0.57 & 0.34 & 0.49 & 0.61 &\textbf{0.87}& 0.52 & 0.54 & 0.52&0.83\\
			DenseNet121 &\checkmark&11&\textbf{0.74}& \textbf{0.58} & 0.45 & 0.45 & \textbf{0.64} &0.86& \textbf{0.57} & \textbf{0.58} & \textbf{0.55}&\textbf{0.88}\\
			EffNetB0 &\checkmark&24&0.71 &0.52 & \textbf{0.46} & 0.44 & 0.61 &0.86& 0.54 & 0.59 & 0.53&0.84\\
			EffNetB7 &\checkmark&21 &0.72 &0.57 & 0.37 & 0.43 & 0.62 &0.85& 0.54 & 0.58 & 0.52&0.85\\
			\hline
		\end{tabular}
    }
\end{table*}

\begin{table*}
	\centering
	\renewcommand{\arraystretch}{1.0}
	\caption{A Comparison of the performance of the top two performance (DenseNet121 and EfficientNet) with data augmentation.}
	\scalebox{1.0}{
	    \label{table:best-results}
		\begin{tabular}{|l|cc|cccc|cc|cccc|}
			\hline
			Method  & \multicolumn{2}{l}{Settings}  &\multicolumn{10}{l|}{Metrics} \\
	               &&& \multicolumn{4}{l}{ Per class F1-Score} & \multicolumn{2}{l}{micro-average} & \multicolumn{4}{l|}{macro-average}\\
			
		 &Pretrained & Best epoch& Control & Infection & Ischaemia & Both & F1 & AUC & Precision & Recall & F1 & AUC\\
			\hline
			\hline
			DenseNet121 &\checkmark&12&0.72& 0.50 & 0.40 & 0.36 & 0.60 &0.86& 0.56 & 0.58 & 0.49&0.85\\
			EffNetB0 &\checkmark&19&0.73 &0.56 & 0.44 & \textbf{0.47} & 0.63 & \textbf{0.87}& \textbf{0.57} & \textbf{0.62} & \textbf{0.55}&\textbf{0.86}\\
			\hline
		\end{tabular}
    }
\end{table*}

\subsection{Performance Metrics} 
The DFUC2021 dataset is imbalanced in terms of its class distribution. To properly handle such class imbalances, the performance is to be reported with per class F1-Score, micro-average F1-Score and area under the Receiver Operating Characteristics Curve (AUC) and macro-average of Precision, Recall, F1-Score and AUC to reflect the overall performance. While the micro-average reflect the overall performance, the macro-average is a good choice in imbalanced multi-class settings \cite{forman2010apples} as it caters well for cases of strong class imbalance. Micro-average aggregates all the True Positives (TP), False Positives (FP), True Negatives (TN) and False Negatives (FN) for all classes to compute the average metrics. This will show the overall performance, but will not reflect if a class with small number of images performed badly. On the other hand, macro-average considers TP, FP, TN and FN for each class $i$ (of n classes), and their respective F1-Scores, as in equation (1). The macro-average F1-Score is determined by averaging the per-class F1-Scores, as in equation (2).
\begin{align}
    \text{macro-average F1-Score} =& \frac{1}{N}\sum_{i}^{N} {F1-Score_i}
\\
   \text{macro-average AUC} =& \frac{1}{N}\sum_{i}^{N} {AUC_i}
\end{align}
where $i={1,..,N}$ represents the $i$-th class and $N$ is the total number of classes, in this case, $N=4$. 
The results will be compared according to macro-average F1-Score. However, for the completeness of scientific assessment, other metrics will be discussed, i.e. macro-average recall (or known as Unweighted Average Recall (UAR)) and macro-average AUC. These metrics provide a balanced judgement on whether an approach can predict all classes equally well, hence reducing the possibility that an approach could be well-fitted to only work for certain classes. 

\section{Experiments}
To benchmark the performance of the networks, we used the input size of $224\times224$ pixels, train the model at learning rate of 0.001 and decrease with a factor of 0.1 if the validation score does not decrease after 5 epochs. For all the models, we set the maximum epochs to 200 and we implemented early stopping if the categorical accuracy of validation does not increase after 8 epochs. We saved the best model of the weights that maximised the score on validation set. We used Keras Tensorflow 2.2.0 on Ubuntu 20.04.2 LTS for our implementation, with 12GB NVIDIA TITAN X GPU and 128GB DDR5 RAM.

\begin{figure}[!htb]
\centering
\includegraphics[width=.12\textwidth]{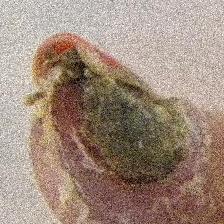}
\includegraphics[width=.12\textwidth]{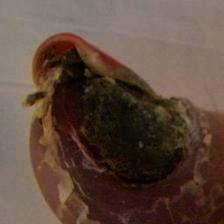}
\includegraphics[width=.12\textwidth]{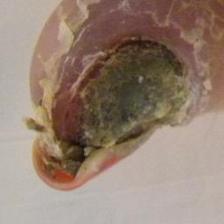}
\includegraphics[width=.12\textwidth]{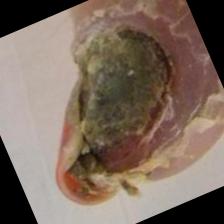}
\caption{Examples of four different data augmentation techniques: noise, contrast, flip vertical and rotate. }
\label{figure:augment}
\end{figure}

The first strategy that we consider is the use of pretrained models from ImageNet. The second strategy is data augmentation, we augment the ischaemia class by using 8 data augmentation techniques, i.e flip horizontal and vertical, rotate -180 degree, add Gaussian noise, crop, shear, scale and adjust contrast by gamma 2.0, yielded a total 2,043 patches after augmentation. We have augment both infection and ischaemia class with 3 data augmentation, i.e. rotate -180 degree, add Gaussian noise and adjust contract by gamma 2.0, yielded a total 2,484 patches after augmentation. Figure \ref{figure:augment} shows some examples of data augmentation.

\begin{table*}
	\centering
	\renewcommand{\arraystretch}{1.0}
	\caption{One-versus-all classification method using EfficientNetB0. Ischaemia-vs-All is a binary classifier for ischaemia vs non-ischaemia; and Infection-vs-All is a binary classifier for infection vs non-infection.}
	\scalebox{1.0}{
	    \label{table:one-vs-all}
		\begin{tabular}{|l|cc|cccc|cc|cccc|}
			\hline
			Method  & \multicolumn{2}{l}{Settings}  &\multicolumn{4}{l|}{Metrics} & \multicolumn{2}{l}{micro-average} & \multicolumn{4}{l|}{macro-average}\\
			
		 &Pretrained & Best epoch& TPR & TNR & FPR & FNR & F1 & AUC & Precision & Recall & F1 & AUC\\
			\hline
			\hline
			Ischaemia-vs-All &\checkmark&16&0.83& 0.93 & 0.17 & 0.068 & 0.92 &0.97& 0.80 & 0.88 & 0.83&0.96\\
			Infection-vs-All &\checkmark&25&0.63 &0.76 & 0.24 & 0.37 & 0.69 &0.77& 0.69 & 0.69 &0.69&0.78\\
			\hline
		\end{tabular}
    }
\end{table*}

\section{Results and Discussion}
We compared the performance of the network with and without pretrained model and observed that the majority of the results are outperformed with pretrained model, as summarised on Table \ref{table:multi-class}. Overall, DenseNet121 performed the best, with macro-average F1-Score of 0.55 and AUC of 0.88. The second top performer is EfficientNetB0, with F1-Score of 0.53 and AUC of 0.84 (better precision and recall compared to InceptionV3). From these results, we selected two best models (on macro-average), DenseNet121 and EfficientNetB0, and conducted further experiments with data augmentation. 

\par
\begin{figure}[!htb]
\centering
\includegraphics[width=.24\textwidth]{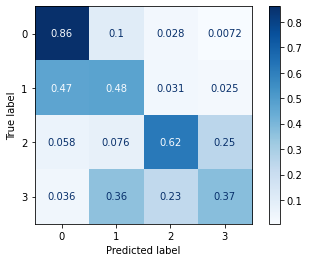}\hfill
\includegraphics[width=.24\textwidth]{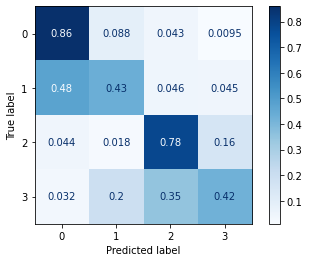}
\caption{Confusion Matrices for the top two best performers: DenseNet121 best result without data augmentation (left) and EfficientNetB0 with data augmentation (right), where class 0 represents control DFU, class 1 represents DFU with infection, class 2 represents DFU with ischaemia and class 3 represents DFU with both conditions.}
\label{figure: confusion}
\end{figure}

\noindent
\textbf{Analysis of the performance of the best models with Data Augmentation}
Table \ref{table:best-results} presents the results of DenseNet121 and EfficientNetB0 with data augmentation. We observed that the performance of DenseNet121 is worse than without data augmentation, particularly on infection. In contrast, EfficientNetB0 improves with balanced classes, which improved in infection F1-Score. Figure \ref{figure: confusion} compares the confusion matrices of DenseNet121 (without data augmentation) and EfficientNetB0 with data augmentation. It is noted that data augmentation improved the recall of ischaemia and both conditions. Overall, EfficientNetB0 achieved the best macro-average of Precision, Recall and F1-Score, with 0.57, 0.62 and 0.55, respectively. The best macro-average AUC of 0.88 is achieved by DenseNet121 (without data augmentation).

\par
\noindent
\textbf{Analysis of the binary classification results using one versus all. }
Table \ref{table:one-vs-all} shows the results of one versus all network, where we achieved high accuracy in F1-Score for ischaemia recognition (0.83) and good result in infection recognition (0.69). These findings is aligned with Goyal et al. \cite{goyal2020recognition} where they have experimented on a smaller dataset.

\par
\begin{figure}[!htb]
\centering
\includegraphics[width=.15\textwidth]{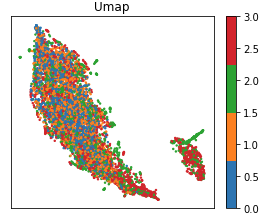}
\includegraphics[width=.15\textwidth]{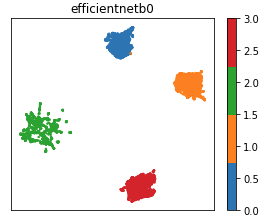}
\includegraphics[width=.15\textwidth]{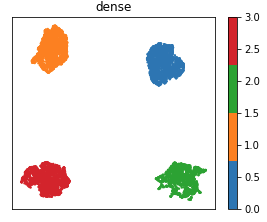} \\
\includegraphics[width=.15\textwidth]{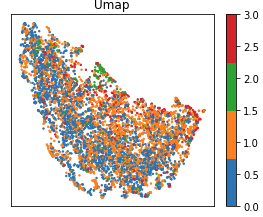}
\includegraphics[width=.15\textwidth]{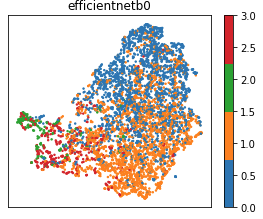}
\includegraphics[width=.15\textwidth]{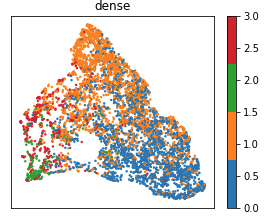}
\caption{UMAP Visualisation of the testing set on the input and the best model (EfficientNetB0) and dense layer.}
\label{figure:umap_bestModel}
\end{figure}

\noindent
\textbf{Analysis of the separability of the best model on training and testing dataset}
We performed statistical analysis on the data distribution separability of the best deep learning algorithms at the final two layers. Figure \ref{figure:umap_bestModel} shows the feature representations produced by UMAP on the training and testing datasets. It visually compared the ability of the CNN networks in clustering the control DFU images (Class 0, blue dots), infection only images (Class 1, orange dots), ischaemic only images (Class 2, green dots) and both conditions (Class 3, red dots). The top row shows the optimal settings for the training set, where the intra-class distance reduced from an average of 4.7248 to 1.9967 and the inter-class distance increased from 7.3041 to 16.6451. However, when tested on unseen testing set as shown in the bottom row, it is noted that the EfficientNetB0 improved the separability of the testing set by reducing the intra-class distance (from an average of 4.5840 to 2.7382) but marginally increase the inter-class distance (from an average of 4.7431 to 4.7794). This analysis demonstrates the challenge in DFUC2021 dataset, particularly the ability of the algorithms to separate infection (orange dots) from control DFU (blue dots), and ischaemia (green dots) from both ischaemia and infection condition (red dots). It is also noted that the dense layer is able to make the distributions more concentrate.

\par
\begin{figure}[!htb]
\centering
\includegraphics[width=.10\textwidth]{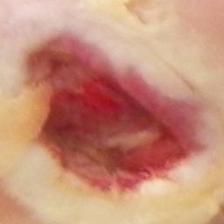}
\includegraphics[width=.10\textwidth]{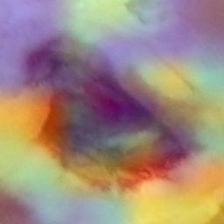}
\includegraphics[width=.10\textwidth]{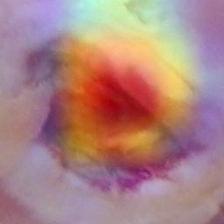} \\
\includegraphics[width=.10\textwidth]{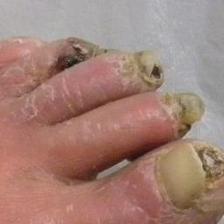}
\includegraphics[width=.10\textwidth]{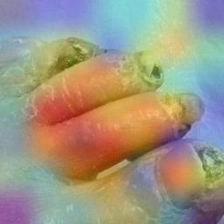}
\includegraphics[width=.10\textwidth]{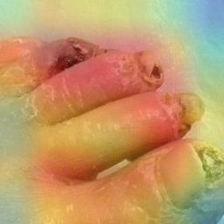}
\caption{Model interpretability. From left to right: Original DFU patches, Grad-CAM visualisation of DenseNet121 and EfficientNetB0. Top row: Example of correctly classified DFU with infection and the corresponding Grad-CAM visualisation \cite{selvaraju2017grad}. The Grad-CAM focuses on the DFU regions. Bottom row: Example of wrongly classified DFU and the corresponding Grad-CAM visualisation. Due to the the networks focus on the wrong regions / background, all the networks predicted DFU with ischaemia as control DFU.}
\label{figure:grad-cam}
\end{figure}

\noindent
\textbf{Model Interpretability}
We performed Grad-CAM visualisation on two cases as shown in Figure \ref{figure:grad-cam}, the top row shows the networks focus on the ulcer region and make correct predictions while the bottom row shows the networks focus with wrong predictions.

\section{Conclusion}
Our experiments show promising results for detection of ischaemia in DFU pathology. However, the detection of infection and detection of co-occurrences of both ischaemia and infection remain a greater challenge. This is due to the difficulties in detecting infection on ``both" categories, as it is often classified as ischaemia, with and without data augmentation. Overall, EfficientNetB0 works best with ImageNet pretrained model and data augmentation. We benchmarked our datasets with some popular deep learning backbones and share the DFUC2021 dataset for reproducibility of the research and encouraging further research to improve the performance of deep learning algorithms in this domain. To take part in DFUC2021 challenge, visit: https://dfu-challenge.github.io for further instructions.

\section*{Acknowledgment}
We gratefully acknowledge the support of NVIDIA Corporation who provided access to GPU resources.


\bibliography{Ref.bib}

\begin{thebibliography}{10}
\providecommand{\url}[1]{#1}
\csname url@samestyle\endcsname
\providecommand{\newblock}{\relax}
\providecommand{\bibinfo}[2]{#2}
\providecommand{\BIBentrySTDinterwordspacing}{\spaceskip=0pt\relax}
\providecommand{\BIBentryALTinterwordstretchfactor}{4}
\providecommand{\BIBentryALTinterwordspacing}{\spaceskip=\fontdimen2\font plus
\BIBentryALTinterwordstretchfactor\fontdimen3\font minus
  \fontdimen4\font\relax}
\providecommand{\BIBforeignlanguage}[2]{{%
\expandafter\ifx\csname l@#1\endcsname\relax
\typeout{** WARNING: IEEEtran.bst: No hyphenation pattern has been}%
\typeout{** loaded for the language `#1'. Using the pattern for}%
\typeout{** the default language instead.}%
\else
\language=\csname l@#1\endcsname
\fi
#2}}
\providecommand{\BIBdecl}{\relax}
\BIBdecl

\bibitem{cho2018idf}
N.~Cho, J.~Shaw, S.~Karuranga, Y.~Huang, J.~da~Rocha~Fernandes, A.~Ohlrogge,
  and B.~Malanda, ``Idf diabetes atlas: Global estimates of diabetes prevalence
  for 2017 and projections for 2045,'' \emph{Diabetes research and clinical
  practice}, vol. 138, pp. 271--281, 2018.

\bibitem{armstrong2017diabetic}
D.~G. Armstrong, A.~J. Boulton, and S.~A. Bus, ``Diabetic foot ulcers and their
  recurrence,'' \emph{New England Journal of Medicine}, vol. 376, no.~24, pp.
  2367--2375, 2017.

\bibitem{yap2018new}
M.~H. Yap, K.~E. Chatwin, C.-C. Ng, C.~A. Abbott, F.~L. Bowling, S.~Rajbhandari
  \emph{et~al.}, ``A new mobile application for standardizing diabetic foot
  images,'' \emph{Journal of diabetes science and technology}, vol.~12, no.~1,
  pp. 169--173, 2018.

\bibitem{goyal2018robust}
M.~{Goyal}, N.~D. {Reeves}, S.~{Rajbhandari}, and M.~H. {Yap}, ``Robust methods
  for real-time diabetic foot ulcer detection and localization on mobile
  devices,'' \emph{IEEE Journal of Biomedical and Health Informatics}, vol.~23,
  no.~4, pp. 1730--1741, July 2019.

\bibitem{goyal2018dfunet}
M.~{Goyal}, N.~D. {Reeves}, A.~K. {Davison}, S.~{Rajbhandari}, J.~{Spragg}, and
  M.~H. {Yap}, ``Dfunet: convolutional neural networks for diabetic foot ulcer
  classification,'' \emph{IEEE Transactions on Emerging Topics in Computational
  Intelligence}, pp. 1--12, 2018.

\bibitem{goyal2020recognition}
M.~Goyal, N.~D. Reeves, S.~Rajbhandari, N.~Ahmad, C.~Wang, and M.~H. Yap,
  ``Recognition of ischaemia and infection in diabetic foot ulcers: Dataset and
  techniques,'' \emph{Computers in Biology and Medicine}, vol. 117, p. 103616,
  2020.

\bibitem{yap2020deep}
M.~H. Yap, R.~Hachiuma, A.~Alavi, R.~Brungel, M.~Goyal, H.~Zhu, B.~Cassidy,
  J.~Ruckert, M.~Olshansky, X.~Huang \emph{et~al.}, ``Deep learning in diabetic
  foot ulcers detection: A comprehensive evaluation,'' \emph{arXiv preprint
  arXiv:2010.03341}, 2020.

\bibitem{mcinnes2018umap}
L.~McInnes, J.~Healy, and J.~Melville, ``Umap: Uniform manifold approximation
  and projection for dimension reduction,'' \emph{arXiv preprint
  arXiv:1802.03426}, 2018.

\bibitem{selvaraju2017grad}
R.~R. Selvaraju, M.~Cogswell, A.~Das, R.~Vedantam, D.~Parikh, and D.~Batra,
  ``Grad-cam: Visual explanations from deep networks via gradient-based
  localization,'' in \emph{Proceedings of the IEEE international conference on
  computer vision}, 2017, pp. 618--626.

\bibitem{cassidy2020dfuc2020}
B.~Cassidy \emph{et~al.}, ``Dfuc2020: Analysis towards diabetic foot ulcer
  detection,'' \emph{arXiv preprint arXiv:2004.11853}, 2020.

\bibitem{tzutalingit}
tzutalin, ``Labelimg,'' https://github.com/tzutalin/labelImg, 2015.

\bibitem{Buchner2020}
J.~Buchner, ``Image hash,'' \url{https://github.com/JohannesBuchner/imagehash},
  2020.

\bibitem{forman2010apples}
G.~Forman and M.~Scholz, ``Apples-to-apples in cross-validation studies:
  pitfalls in classifier performance measurement,'' \emph{Acm Sigkdd
  Explorations Newsletter}, vol.~12, no.~1, pp. 49--57, 2010.

\end{thebibliography}
\bibliographystyle{IEEEtran}




\end{document}